\pdfoutput=1
\documentclass[11pt]{article}

\usepackage{acl}

\usepackage{times}
\usepackage{latexsym}
\usepackage{hyperref}
\usepackage{booktabs}

\usepackage[T1]{fontenc}
\usepackage[utf8]{inputenc}

\usepackage{microtype}

\usepackage{url}

\usepackage{xcolor}
\newcommand{\oov}[1]{\textcolor{red}{\textit{#1}}}
\newcommand{\mightbe}[1]{\textcolor{blue}{\textit{#1}}}
\usepackage{tabularx}
\usepackage{arydshln}
\usepackage{makecell}
\usepackage{multicol}
\usepackage{multirow}
\usepackage{tikz}
\usetikzlibrary{pgfplots.groupplots}
\usepackage{pgfplots}\pgfplotsset{width=7.5cm,compat=1.17}
\usepackage{placeins}
\usepackage{subcaption}
\usepackage{stfloats }

\usepackage{todonotes}

\title{The University of Edinburgh's Submission to the WMT22 Code-Mixing Shared Task (MixMT)}

\author{Faheem Kirefu \qquad Vivek Iyer \qquad Pinzhen Chen \qquad  Laurie Burchell \\
School of Informatics, University of Edinburgh \\
\texttt{\{fkirefu,vivek.iyer,pinzhen.chen,laurie.burchell\}@ed.ac.uk}}

\begin{document}
\maketitle
\begin{abstract}

The University of Edinburgh participated in the WMT22 shared task on code-mixed translation. This consists of two subtasks: i) generating code-mixed Hindi/English (Hinglish) text generation from parallel Hindi and English sentences and ii) machine translation from Hinglish to English. As both subtasks are considered low-resource, we focused our efforts on careful data generation and curation, especially the use of backtranslation from monolingual resources. For subtask 1 we explored the effects of constrained decoding on English and transliterated subwords in order to produce Hinglish. For subtask 2, we investigated different pretraining techniques, namely comparing simple initialisation from existing machine translation models and aligned augmentation. For both subtasks, we found that our baseline systems worked best. Our systems for both subtasks were one of the overall top-performing submissions.
% please all words should be capitalised in \section{} 
% but only the first word should be capitalised in \sub*section{}

\end{abstract}

\section{Introduction}

Code-mixing is the shift from one language to another within a single conversation or utterance \citep{sitaram2019survey}. It is an extremely common and diverse communicative phenomenon worldwide \citep{dogruoz-etal-2021-survey, sitaram2019survey}, though one which is currently under-served by many NLP technologies \citep{calcs-2021-approaches}. 

One of the most well-known examples of code-mixing is between Hindi and English, commonly referred to as Hinglish\footnote{In the scope of this paper, we designate ``hg'' as the language code for Hinglish.}. It is extremely common amongst Hindi-English bilingual speakers in both speech and text, used across a range of genres and media \citep{parshad2016india}, and has its own distinctive features and linguistic forms \citep{kumar1986-ru,sailaja2011-js}. The process of generating Hinglish from the written text is non-trivial, as code-mixing may happen at the phrase or word level, but Hindi and English differ substantially syntactically.

As a novel addition to the current code-mixing NLP research, we investigated lexically constraining the Hinglish output in subtask 1 to only contain words from English and Hindi sources. Through analysis, we demonstrated that transliteration mismatches could affect performance.

% Paragraph on Aligned Augmentation (Vivek)
Another novel approach we explore for this task, particularly for subtask 2, is a denoising-based pretraining technique called Aligned Augmentation (AA) \cite{pan-etal-2021-contrastive}. AA, which trains MT models to denoise artificially generated code-mixed text, was shown by \citet{pan-etal-2021-contrastive} to boost translation performance across a variety of languages - thanks to the enhanced transfer learning brought about by code-mixed pretraining. In this work, we explored if this general-purpose approach could be useful for translating authentic, human-generated code-mixed text, focusing on Hinglish.
% Due to the similarity between the training and the test data, we hypothesised that such a model could adapt well to Hinglish-English translation. %More importantly, this approach allowed us to leverage abundant, high-quality monolingual data in English, Hindi and Hinglish and could help in learning meaningful representations across Hinglish, English and Hindi word translations to robustly translate noisy data.

Despite these efforts, we found that for both subtasks our original baselines worked better and constituted our final submissions for this task, which ranked as one of the top-performing systems for both subtasks, by both automatic and human evaluation. We hope our methods, particularly Hinglish data generation, that allowed us to build these systems would be useful to the community; as would the findings from our additional research explorations.

\section{Related Work}

\subsection{Code-mixing}
Due to an increasing prevalence of code-mixed data on the Internet, there is a growing body of research into code-mixing, particularly for Hinglish, in the NLP community. \citet{dogruoz-etal-2021-survey} provide a comprehensive literature review of code-mixing in the context of language technologies. Whilst they highlight several challenges inherent in NLP with code-mixed text (such as understanding cultural and linguistic context, evaluation, and a lack of user-facing applications), the most notable obstacle for this shared task is the lack of data. They note that there are very few code-mixed datasets, making it challenging to build deep learning models such as those for NMT. In this work, we use backtranslation as our main data augmentation method \citep[][\textit{inter alia}]{edunov-etal-2020-evaluation,barrault-etal-2020-findings,akhbardeh-etal-2021-findings}. This allows us to leverage the larger amount of monolingual data for better final model performance. The XLM toolkit \citep{lample2019cross} seemed an ideal choice to backtranslate our Hinglish. This is because it has shown promising results in unsupervised and semi-supervised settings where parallel data is sparse, but monolingual data is ample. Also given that Hinglish is closely related to both languages, we believed Hinglish should be an ideal language to use in a semi-supervised setting.

\subsection{Constrained decoding}
Constrained decoding involves applying restrictions to the generation of output tokens during inference. Most implementations have the goal of ensuring that desired vocabulary items appear in the target side sequence \citep{hokamp-liu-2017-lexically, hasler-etal-2018-neural, post-vilar-2018-fast}. Alternatively, \citet{kajiwara-2019-negative} paraphrase an input sentence by forcing the output to not include source words, and \citet{chen-etal-2020-parallel-sentence} constrain NMT decoding to follow a corpus built in a trie data structure to find parallel sentences.

To the best of our knowledge, previous linguistics research investigated and applied the grammatical constraints in code-mixing \citep{government-constraint,functional-head-constraint,inversion-constraint}, rather than the novel method in our work of introducing lexical constraints.

\subsection{Aligned augmentation}
Several recent works \citep{Yang_Ma_Zhang_Wu_Li_Zhou_2020, yang-etal-2020-csp, lin-etal-2020-pre, pan-etal-2021-contrastive} have explored enhancing cross-lingual transfer learning by pretraining models on the task of `denoising' artificially code-mixed text. Methods to create the necessary code-mixed data vary, and include bilingual or multilingual datasets and word aligners \citep{Yang_Ma_Zhang_Wu_Li_Zhou_2020, yang-etal-2021-multilingual}, lexicons \citep{yang-etal-2020-csp, lin-etal-2020-pre, pan-etal-2021-contrastive}, or combining code-mixed noising with traditional masked noising approaches \citep{lietal-2022-cemat}.

The most successful among these methods is Aligned Augmentation (AA) \cite{pan-etal-2021-contrastive}, which randomly substituting words in the source sentence with their word-level translations, as obtained from a MUSE \cite{lample2018word} dictionary. \citet{pan-etal-2021-contrastive} showed that their technique can effectively align multilingual semantic word representations and boost performance across various languages. However, these methods focus on training general-purpose MT models. In this work, we investigate their utility for translating real human-generated code-mixed text.

\subsection{Automatic evaluation metrics}

Automatic translation evaluation is usually done using BLEU \citep{papineni-etal-2002-bleu}, yet there is no comprehensive study on its suitability for code-switched translation. Specifically in this task, the organisers announced that the participating systems will be evaluated using ROUGE-L \citep{lin-2004-rouge} and word error rate (WER). Nonetheless, the packages implementing these metrics were not specified. Since ROUGE comes with different language, stemming and tokenisation settings, we instead used BLEU, ChrF++ \citep{popovic-2017-chrf}, translation error rate (TER), and WER\footnote{\url{https://github.com/jitsi/jiwer}} for our internal validation. The first three are as implemented with sacreBLEU \citep{post-2018-call}. We stick to the default configurations, except that the ChrF word n-gram order is explicitly set to 2 to make it ChrF++. In addition, the organisers performed a small-scale human evaluation on 20 test instances for all submissions.

In this work, we advocate for a character-based metric when evaluating the Hinglish output in subtask 1. This is because for the code-switched language, there is no formal spelling or defined grammar, and words may have a diverse range of acceptable transliterations and lexical forms.

\section{Subtask 1: Translating into Hinglish}

Good quality Hinglish data is hard to come by, and parallel Hinglish data with Hindi or English even more frugal. Therefore, for both subtasks we concentrated our efforts on generating good Hinglish backtranslation. We planned to use the model which produced the highest quality Hinglish for subtask 1 as our backtranslator for subtask 2, hence we focused our efforts on each subtask sequentially.

\subsection{Data cleaning and preprocessing}

% Added by Laurie
After deduplicating the data, we removed non-printing characters and normalised the punctuation. We then ran rule-based filters, removing any sentences with fewer than two or more than 150 words, where fewer than 40\% of the words are written in the relevant script, or where over 50\% of characters are not letters in the relevant script. For English and Hindi, we ran \texttt{fasttext} language ID and removed any sentence which was not classified as the relevant language.\footnote{Our cleaning scripts are adapted from those provided by the Bergamot project. \href{https://github.com/browsermt/students/tree/master/train-student}{https://github.com/browsermt/students/tree/master/train-student} Specifically, we add support for Hindi and Hinglish text.} For Hinglish, we also removed any sentence with a predicted probability of English greater than 0.99 in order to remove sentences that were solely in English. We tokenised English and Hinglish using Moses scripts \citep{DBLP:conf/acl/KoehnHBCFBCSMZDBCH07} and tokenised Hindi using the \texttt{indicnlp} library \citep{kunchukuttan2020indicnlp}.

We decided to add explicit preprocessing and postprocessing capabilities for handling social media text, given that this was the domain for subtask 2. On both source and target sides, we replaced URLs, Twitter handles, hashtags and emoticons each with their own placeholder tokens\footnote{\texttt{<URL>, <TH>, <HT>} and \texttt{<EMO>} respectively}, to be replaced back from the source after inference. These placeholders made up 1.7\% of the validation set tokens for subtask 2, far higher than would appear in general domain data.

\label{aa:data-subtask2}
\subsubsection{The HinGe dataset}

The primary dataset for subtask 1 was the HinGe dataset \citep{srivastava-singh-2021-hinge}, which consisted of hi-en-hg parallel sentences, with some examples synthetic and some human-generated. This was provided to us pre-split into training and development sets for both data types. However, we noticed that these sets were not mutually exclusive, and after deduplication and filtering on the synthetic data human annotations\footnote{We only kept sentences with an average rating greater than 4, and annotator disagreement less than 5}, we had 6,727 hi-en-hg examples in total.

\subsubsection{Base hi$\leftrightarrow$en translation models}

Firstly, we trained four Transformer-base  \citep{vaswani2017attention} models with different seeds using Marian \citep{mariannmt} for both hi$\rightarrow$en and en$\rightarrow$hi directions, using the data from the hi-en parallel Samanantar corpus\footnote{Each sentence was annotated with the LaBSE \citep{feng-etal-2022-language} Alignment Score (between 0 and 1), so we filtered out values less than 0.65,  resulting in around 10.1M sentences} \citep{ramesh2021samanantar}. Given the findings of \citet{DBLP:conf/mtsummit/DingRD19} with regard to vocabulary choice for low-resource scenarios, and that our task inherently contains transliteration, we opted for a low BPE \citep{sennrich-etal-2016-neural} merge size of 4k, resulting in a small joint vocabulary of 7.9k. We used the hi-en FLORES development set  \citep{GoyalGCCWJKRGF22} for validation and early stopping, and noticed our model produced surprisingly good quality translations in both directions\footnote{sacreBLEU: 33.8 for hi$\rightarrow$en and 32.7 for hi$\rightarrow$en on FLORES development set}. We used these models (along with vocabulary) to both initialise subsequent models and generate backtranslation for more training data.

\subsubsection{Hinglish data}

L3Cube-HingCorpus \citep{nayak2022l3cube} and CC-100 Hindi Romanized \citep{DBLP:conf/acl/ConneauKGCWGGOZ20} are two Hinglish corpora that we wished to backtranslate into both English and Hindi. Given that we only had a small amount of parallel Hinglish data, compared to our `monolingual' datasets, we used the XLM toolkit \citep{lample2019cross} to train a semi-supervised model (see Appendix~\ref{sec:xlmrdeets} for details). We then backtranslated the monolingual Hinglish data into both English and Hindi. However, given the noisy quality of the data and translations themselves, we decided to evaluate them using our hi$\rightarrow$en and en$\rightarrow$hi Marian models. Specifically, for an en-hi backtranslated (XLM) sentence pair, we translated the en/hi into hi/en respectively, then evaluated the double translated output using ChrF, with the XLM backtranslations as the references. We then took a mean of the English and Hindi ChrF score to get our final confidence value. We used the resulting hg-en-hi sentence trios with values at least 0.4, to compromise between the quality and quantity of data available to use as training. Most of the sentences scored quite poorly, and filtering on 0.4 yielded 2.1M sentences, only about 12\% of the original Hinglish monolingual dataset.

\begin{table*}[bht]
\centering
\begin{tabular}{@{}lcccc@{}}
\toprule
Beam Size & BLEU ($\uparrow$) & ChrF++ ($\uparrow$) & TER ($\downarrow$) & WER ($\downarrow$) \\
\midrule
\multicolumn{5}{l}{\textit{Unconstrained}} \\
1  & 17.8 & 42.8 & 65.3 & \textbf{81.5} \\
4  & \textbf{18.1} & \textbf{44.0} & \textbf{64.5} & 85.7 \\
12 & 18.0 & 43.8 & 64.8 & 86.0 \\
24 & 18.0 & 43.7 & 65.0 & 85.5 \\
36 & 17.9 & 43.5 & 65.1 & 85.4 \\
48 & 18.0 & 43.6 & 65.2 & 85.5 \\
\midrule
\multicolumn{5}{l}{\textit{Constrained}} \\
1  & 10.8 & 33.1 & 76.1 & 75.1 \\
2  & 12.2 & 35.6 & 74.9 & 69.1 \\
4  & 13.2 & 36.6 & 74.2 & 63.5 \\
6  & 14.1 & 37.7 & 73.5 & 60.8 \\
12 & 14.6 & 38.1 & 73.7 & 58.6 \\
24 & 14.8 & 38.5 & 73.5 & 57.2 \\
36 & 14.9 & 38.7 & 73.6 & \textbf{56.7} \\
48 & 15.0 & 38.7 & 73.6 & 57.0 \\
\bottomrule
\end{tabular}
\caption{Experimental results on the validation set with unconstrained and constrained decoding for subtask 1.}
\label{tab:constrained-results}
\end{table*}

\subsubsection{Transliteration} \label{subsec:translitermodel}%this is referred to later, so please keep the label if you are re-arranging stuff.

In order to best leverage the Samanantar hi-en parallel corpus, we transliterated the Hindi side into Roman script\footnote{In the scope of this paper, we use ``ht'' to denote pure romanised Hindi transliteration}, on the word level. Although this forward transliteration was not likely to contain much code-mixed text, it would still be useful training data for our model, given that both the Hindi and English sources are assumed to be either the original sources or human translationese.

We used the AI4Bharat Indic transliterator \citep{Madhani2022AksharantarTB}, to convert (on the word level) all romanised tokens contained in our monolingual Hinglish datasets into Devanagari script. This tool is a neural-based model with beam search capabilities, therefore we generated the top 4 results in Hindi for each Hinglish token. We used the top 4 instead of the most likely candidate as, upon inspection, we found that the correct corresponding Hindi token was not always predicted first. We also used a human-generated list of Hinglish-English pairs form the Xlit-Crowd corpus \citep{DBLP:conf/lrec/KhapraRKVB14} which we treated as the gold standard. 

To summarise, our training data for our hi$\rightarrow$ht transliterator\footnote{We decided to build our own transliterator as we found existing tools in this direction to be of poor quality} consists of 5.3M Hinglish-Hindi word pairs (1.3M unique Hinglish words), and 15k from Xlit\-Crowd, of which we use 1k as a validation set for early stopping. We train a small transformer model with Marian on the \textbf{character-level} for both input and output. 
When forward transliterating the Hindi side of the Samanantar corpus, we copied over non-standard strings (such as numbers, punctuation etc.), or else we looked up the token (if it existed) in our gold standard list. Otherwise, we used our transliteration model as a final back-off. In hindsight, one disadvantage of our approach was that we did not generate multiple candidates for each Hindi word, to reflect the diversity of possible romanised candidate tokens. 

We also used this transliteration model as part of our constrained decoding experiments later (see Section \ref{subsec:consdec}).

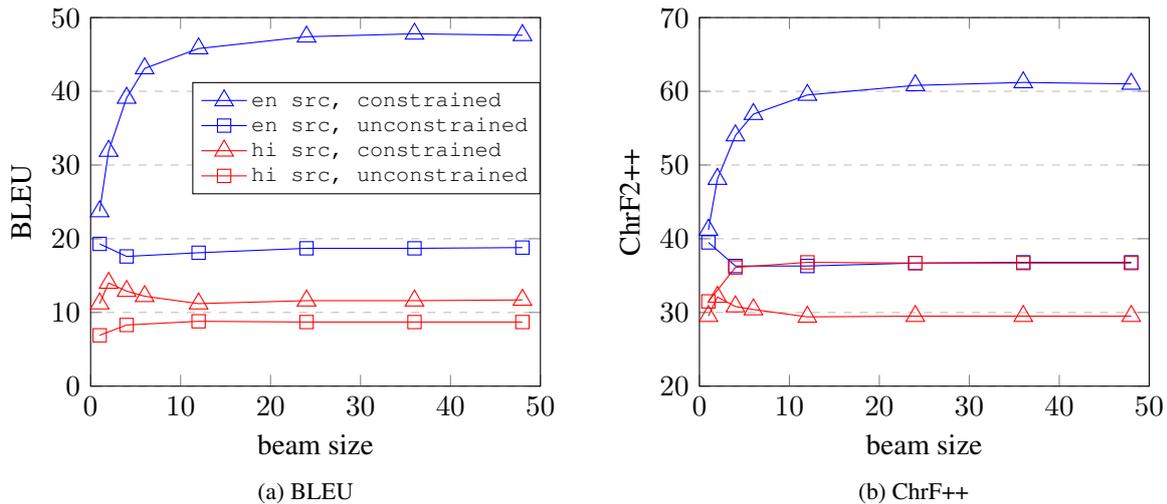
\begin{figure*}[thb]
\begin{subfigure}{.5\textwidth}
    \begin{tikzpicture}
\begin{axis}[
    xlabel={beam size},
    ylabel={BLEU},
    xmin=0, xmax=50,
    ymin=0, ymax=50,
    xtick={0,10,20,30,40,50},
    ytick={0,10,20,30,40,50},
    legend cell align={left},
    legend style={nodes={scale=0.75},at={(1,0.825)}},
    ymajorgrids=true,
    grid style=dashed,
]
\addplot[
    color=blue,
    mark=triangle,
    mark size=4pt
    ]
    coordinates {
    (1,23.7)
    (2,31.9)
    (4,39.1)
    (6,43.1)
    (12,45.8)
    (24,47.4)
    (36,47.8)
    (48,47.6)
    };
    \addlegendentry{\tt en src, constrained}
\addplot[
    color=blue,
    mark=square,
    mark size=2.5pt
    ]
    coordinates {
    (1,19.3)
    (4,17.6)
    (12,18.1)
    (24,18.7)
    (36,18.7)
    (48,18.8)
    };
    \addlegendentry{\tt en src, unconstrained}
\addplot[
    color=red,
    mark=triangle,
    mark size=4pt
    ]
    coordinates {
    (1,11.2)
    (2,14.0)
    (4,12.9)
    (6,12.2)
    (12,11.2)
    (24,11.6)
    (36,11.6)
    (48,11.7)
    };
    \addlegendentry{\tt hi src, constrained}
\addplot[
    color=red,
    mark=square,
    mark size=2.5pt
    ]
    coordinates {
    (1,6.9)
    (4,8.3)
    (12,8.8)
    (24,8.7)
    (36,8.7)
    (48,8.7)
    };
    \addlegendentry{\tt hi src, unconstrained} 
\end{axis}
\end{tikzpicture}
    \caption{BLEU}
    \label{fig:constrained-bleu}
\end{subfigure}%
\begin{subfigure}{.5\textwidth}
    \begin{tikzpicture}
\begin{axis}[
    xlabel={beam size},
    ylabel={ChrF2++},
    xmin=0, xmax=50,
    ymin=20, ymax=70,
    xtick={0,10,20,30,40,50},
    ytick={20,30,40,50,60,70},
    ymajorgrids=true,
    grid style=dashed,
]
\addplot[
    color=blue,
    mark=triangle,
    mark size=4pt
    ]
    coordinates {
    (1,41.2)
    (2,48.1)
    (4,54.0)
    (6,56.9)
    (12,59.5)
    (24,60.8)
    (36,61.2)
    (48,61.0)
    };
\addplot[
    color=blue,
    mark=square,
    mark size=2.5pt
    ]
    coordinates {
    (1,39.5)
    (4,36.3)
    (12,36.3)
    (24,36.7)
    (36,36.8)
    (48,36.8)
    };
\addplot[
    color=red,
    mark=triangle,
    mark size=4pt
    ]
    coordinates {
    (1,29.5)
    (2,32.1)
    (4,30.8)
    (6,30.4)
    (12,29.4)
    (24,29.5)
    (36,29.5)
    (48,29.5)
    };
\addplot[
    color=red,
    mark=square,
    mark size=2.5pt
    ]
    coordinates {
    (1,31.5)
    (4,36.1)
    (12,36.8)
    (24,36.7)
    (36,36.7)
    (48,36.7)
    }; 
\end{axis}
\end{tikzpicture}
    \caption{ChrF++}
    \label{fig:constrained-chrf2++}
\end{subfigure}
\caption{Validation BLEU and ChrF++ of the constrained and unconstrained outputs scored against English and transliterated Hindi \textit{sources} separately.}
\end{figure*}

\subsection{Baseline (unconstrained decoding)}

We decided to use a dual encoder setting given that we have two inputs in this task, and initialise our model from our previously trained Marian MT systems. We used hi$\rightarrow$en to initialise the Hindi-decoder and the English-encoder cross attention parameters, whereas en$\rightarrow$hi was used to initialise the English-encoder and all other decoder parameters. Our vocabulary was the same as the pretrained models.

Early stopping with patience 10 on the HinGe dataset was used for convergence - for all of the experiments mentioned in this paper. Our training regime consisted of two stages:

\begin{itemize}
    \item General domain - The training datasets used were the backtranslated Hinglish and forward transliterated Samanantar corpora. We used all of the HinGe dataset as a validation set.
    \item Finetuning - We continue training on a subset of HinGe dataset, using a distinct smaller subset (1k) of it as a validation set.
\end{itemize}

\subsection{Constrained decoding} \label{subsec:consdec}

After analysing the training data, we hypothesized that nearly all the output words should either be from the English source, or as a transliteration of a word from the Hindi source, with likely little change in sentence structure. This inspired us to use the technique of constrained decoding when generating Hinglish. 

Unlike standard constrained decoding where a model is forced to incorporate certain words in the output, our proposal is to exclude vocabulary words that do not exist in English or transliterated Hindi source sentences. Following \citet{chen-etal-2020-parallel-sentence}'s notion, we applied pre-expansion pruning: disallowed word paths are assigned an extremely small score before hypotheses are ranked and expanded. Specifically, to obtain Hindi transliteration, we used our transliteration model described in Section \ref{subsec:translitermodel}.

We performed beam searches with constrained decoding and reported automatic scores on the validation set in Table~\ref{tab:constrained-results}. Unfortunately, constrained decoding does not beat unconstrained decoding. As a general trend, WER and TER do not change much as beam size increases, while BLEU and ChrF++ significantly improve.

To better understand the impact of constrained decoding, we score the validation outputs against English and transliterated Hindi sources separately, then plot BLEU and ChrF++ numbers in Figure~\ref{fig:constrained-bleu} and Figure~\ref{fig:constrained-chrf2++}. We observe that with increasing beam sizes, constrained decoding prefers to generate English tokens instead of transliterated Hindi. Unconstrained decoding achieves a much better balance.

One hypothesis is that the quality of Hindi transliteration is not perfect, resulting in the model preferring English tokens from the vocabulary. Hence, we compute the percentage of words in the gold reference as well as in the unconstrained (baseline) output that come from neither the English nor the transliterated Hindi source. Surprisingly, on average 45.1\% of the total words in the unconstrained output do not appear in the sources; as for the gold reference, it is 39.8\% which is slightly lower. It is worth noting that the numbers might be inflated as we computed the word overlap after outputs are detokenised. Yet it implies that many of the reference words do not exactly appear in the lexical constraints determined from the source senteneces.

Finally, we visualise the first five validation sentences in Table~\ref{tab:constrained-analysis}. We highlight in \oov{red} the target words that do not exist in the source sentences; we also label the possible corresponding tokens from the sources in \mightbe{blue}. It can be confirmed that most mismatches are due to differences in Hindi transliteration and letter cases. This indicates that the lexical constraint idea is suitable in theory, but it is hindered by the error propagation in transliteration. This may have been alleviated by running multiple transliteration schemes on the Hindi source to make the constraints more diversified.

\begin{table*}[bht]
\centering
% \small
% \begin{tabular}{lp{120mm}}
% \toprule
% \input{analysis-constrained/examples.tex}
\includegraphics[trim={70 390 70 70},clip]{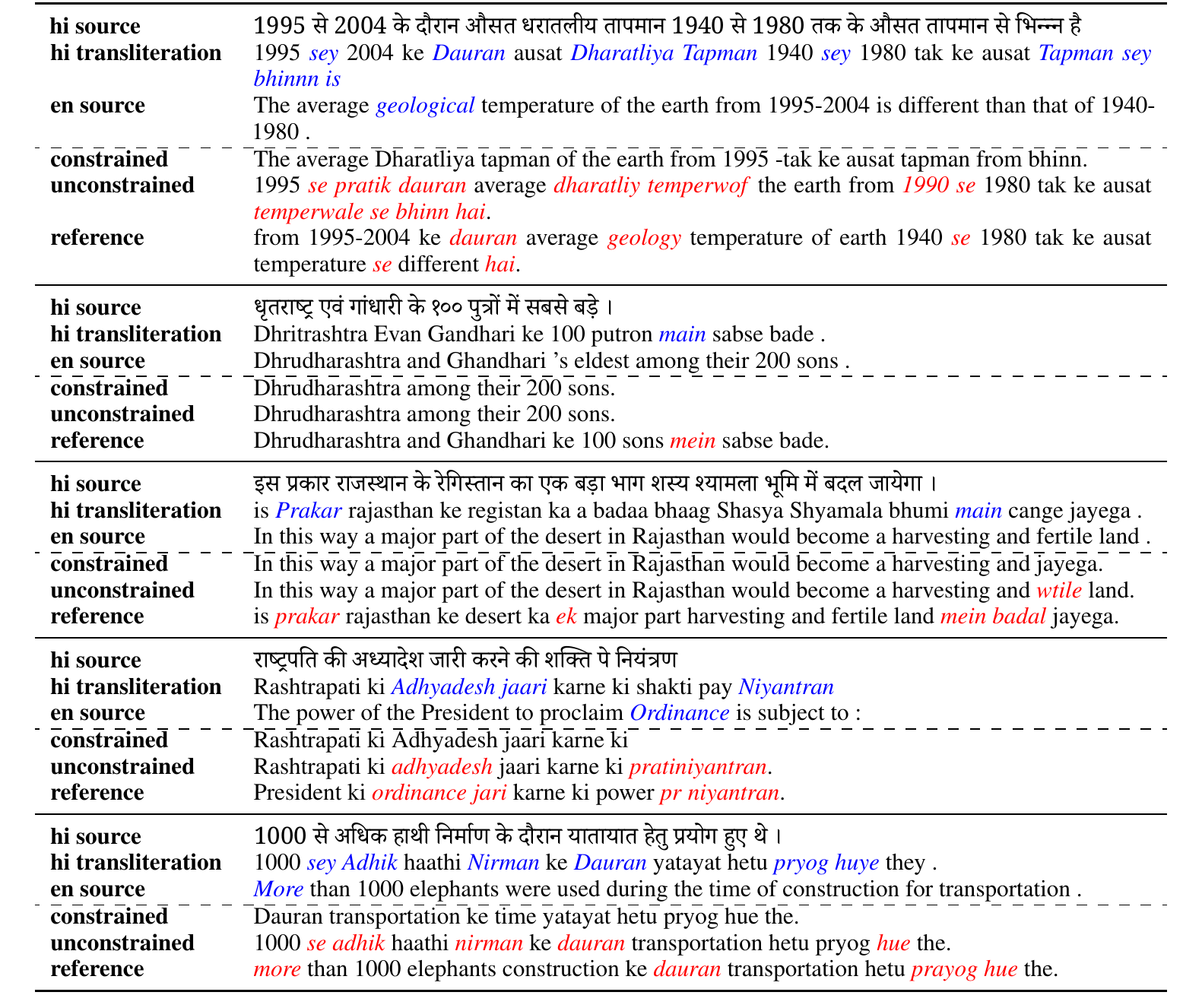}
% \end{tabular}
\caption{The first five validation instances: English and Hindi sources, as well as constrained, unconstrained and reference outputs. \oov{red} denotes the target side words that do not appear in either of the source sentences from a constrained aspect; \mightbe{blue} denotes possible source-target matches in a different surface form.}
\label{tab:constrained-analysis}
\end{table*}

\section{Subtask 2: Hinglish-to-English}

\subsection{Data cleaning and preprocessing}

The primary dataset provided for this task PHINC \citep{srivastava-singh-2020-phinc} is relatively small at 13.7k English-Hinglish pairs. Therefore, we aimed to generate domain-specific parallel data using our  baseline model from Subtask 1 on English monolingual data.

We analysed the source side of the validation dataset to determine the most frequent content words (see Table \ref{tab:hatewords}) and then selected these words (and any morphological/spelling variants) from the English WikiMatrix corpus \citep{schwenk-etal-2021-wikimatrix}. This yielded a total 477k English sentences and we henceforth refer to this selection of sentences as ToxicWiki. We also used Sentiment140 \citep{DBLP:conf/comsnets/SahniCCS17}, a dataset of 1.6M tweets in English, as the domain of our validation set is also Twitter.

\begin{table}[hbt!]
    \centering
    \begin{tabular}{lrr}
    \toprule
     Word & Validation & WikiMatrix   \\
     \midrule
      rape   & 249 & 23,198 \\
      hate   & 117 & 16,824 \\
      terrorism   & 24 &  11,160 \\
      khoon (blood)  & 21 & 59,526  \\
      murder  & 21 & 75,066   \\
      india   & 16 & 291,054 \\
     \midrule

      \textbf{Total} & - & 476,828 \\
    \bottomrule
    \end{tabular}
    \caption{Frequency of top content words present in our validation set, and the number of sentences within WikiMatrix that contained the word (or morphological variants). The resulting sentences formed ToxicWiki}
    \label{tab:hatewords}
\end{table}

To obtain the Hinglish side of both Sentiment140 and ToxicWiki datasets, we backtranslated into Hindi using our en$\rightarrow$hi Marian model, and then used the en-hi pair and our baseline system for subtask 1 to obtain the corresponding Hinglish. However, many of the placeholders (such as \texttt{<HT>}) did not occur frequently enough during the training of subtask 1 for the model to learn to consistently copy them across; therefore the model was not able to predict them with a large degree of accuracy. Therefore, we ran a postprocessing script that corrected for placeholders on the backtranslated Hinglish, given the English source, so that our downstream model would be able to learn to simply copy these placeholders across. Specifically, we made sure that the number of each placeholder type in the backtranslated Hinglish was the same (and in roughly the same position) as that in the source sentence.

For the AA experiments described in Section \ref{subsec:aa-subtask2}, we used monolingual Hindi, English and Hinglish data. For Hindi and English, we randomly sampled 20M sentences from the News Crawl corpora \citep{akhbardeh-etal-2021-findings}. For Hinglish, the monolingual corpora described above was used. In order to code-mix these corpora as described in the AA algorithm, we used MUSE dictionaries for the Hindi-English pair. For Hinglish-Hindi pairs, we used the data generated with AA for the transliteration model.

\subsection{Baseline systems}

We used a hi$\rightarrow$en MT to initialise the baseline hg$\rightarrow$en model.

Our training regime consisted of three stages:

\begin{enumerate}
    \item General - Training on the backtranslated en-hg internet corpora (with confidence value at least 0.4), and ht-en side of the Samanantar corpus, where we treat the transliteration as Hinglish. We used the PHINC dataset as our validation set for early stopping.
    \item We continued training on Sentiment140 and ToxicWiki corpus, using the same validation set as before, until convergence.
    \item We continued training on the PHINC dataset, using a small subset (1k) of it as validation data for early stopping.
    
\end{enumerate}

As we had multiple hi$\rightarrow$en MT systems, we also trained an ensemble model of four, where we followed the same training regime above with parameters initialised from each of our  hi$\rightarrow$en models. Our results are shown in Table \ref{tab:subtwobaseresults}, with our ensemble model outperforming the single on all metrics.

\begin{table*}[!th]
\centering
\begin{tabular}{@{}lcccc@{}}
\toprule
 & BLEU ($\uparrow$) & ChrF++ ($\uparrow$) & TER ($\downarrow$) & WER ($\downarrow$) \\ \midrule
\multicolumn{5}{l}{\textit{Baseline Experiments}} \\
Single model & 24.5 & 47.0 & 65.1 & 72.0 \\ 
Ensemble (of 4) & \textbf{25.5} & \textbf{48.7} & \textbf{62.9} & \textbf{70.5} \\
\bottomrule
\end{tabular}
\caption{Baseline results for subtask 2 on the MixMT validation set.}
\label{tab:subtwobaseresults}
\end{table*}

\subsection{Aligned Augmentation for subtask 2}
\label{subsec:aa-subtask2}

Our Aligned Augmentation (AA) experiments where implemented with Fairseq \citep{ott2019fairseq}, and we used the Transformer architecture, with 12 encoder and 12 decoder layers. Our first step consisted of pretraining these models on Hindi, English, and Hinglish corpora, with the target being the ``denoised'' sentence - thus training the model to reconstruct the original sentence, following the AA algorithm. For validation, we randomly sampled 1k sentences from the training corpus. 

We then finetuned this model on the Hinglish-English parallel corpora mentioned above. The major AA baselines we trained and their performances are listed in Table \ref{tab:aa-results} - along with a randomly initialised baseline that was trained solely on the parallel corpora. The data sources we used in our experiments were quite diverse: we started with high-quality monolingual data for pretraining followed by parallel datasets of varying domains and qualities, (the Hinglish backtranslated corpora, Sentiment140, PHINC and ToxicWiki). We attempted to explore how best these resources could be utilised. We started with our default training paradigm: we finetuned on backtranslated Hinglish, followed by the ToxicWiki and then a shuffled concatenation of the social media datasets - the Sentiment140 and PHINC datasets respectively. This was based on the intuition that the final model should be most recently trained on datasets from similar domains as the test set.

Following this paradigm, we conducted two sets of experiments: a ``validation experiment'' that tries to estimate the best choice of validation sets, and ``training experiments'' to verify the importance of some training sources empirically. The former is a crucial decision in our experiments given our use of early stopping. We find that validating on the official MixMT validation sets released for Subtask 2 ends up performing significantly worse than validating on a subset of the respective training datasets. This is surprising given the performances reported in Table \ref{tab:aa-results} are evaluated on the same validation sets. This suggests that training and validating the model on corpora from different domains can help boost the final performance - even if it does not improve loss on the final validation set. In the latter body of experiments, we attempted to determine the value of the XLM backtranslated corpora on performance - which seems very noisy on manual inspection, with the target side (English) being generated through backtranslation. Surprisingly, its inclusion significantly enhances performance, by +5 BLEU points. This could be due to various reasons: its sheer size (15M sentences), the presence of word-level translations between English to Hinglish in parallel sentences (despite grammatical errors), the similarity between the source and the target encouraging ``copying" which can sometimes be beneficial for this task, etc. We also find that the inclusion of hi-en along with hg-en further boosts performance, consistent with the findings of previous works on multilingual MT. We empirically found that including `all' available hi-en sentences and `all' available hg-en sentences was more beneficial than splitting our parallel dataset into the two respective directions -- despite the target sentence being duplicated in the former.

Compared to the Random baselines, our final AA baselines show consistent improvement for all given metrics - though the improvement is not very significant with respect to BLEU o TER. A closer glance at the validation set and the generated predictions reveals the potential reason behind this - there is a significant amount of noise present in the validation sets due to the social media domain, with errors in both syntax and semantics. Given that it is not always easy to comprehend and translate such sentences well, the gold reference sentences are sometimes of relatively poor quality - containing various potential errors such as inaccurate word form predictions, grammatical errors, misspellings etc. While word-based metrics may fail to handle these cases; ChrF++, being a character-based metric, can likely alleviate noise that may have propagated to reference sentences and might be a more suitable metric for Subtask 2 as well. It is encouraging to note AA's improvement over the Random baseline in this light.

AA appears to bring about some improvement qualitatively, especially regarding noisy input - for instance, it was able to more accurately translate misspellings and handle grammatical inconsistencies. However, the frequency of sentences where AA performs better than its randomly initialized counterparts seems relatively low. One explanation could be that fine-tuning the model on 18M parallel sentences could lead it to `forget' the representations learned during pretraining. This is in line with the findings of \cite{pan-etal-2021-contrastive} that observe relatively lower improvements for high-resource languages. While adding large corpora (15M sentences) such as the XLM backtranslated corpora does lead to net improvements, it is possible optimization in the size of finetuning data used could lead to even greater gains. Secondly, given that our dictionaries appear to help in noise resolution, it might be useful to incorporate various types of misspellings rigorously in the code-mixing lexicons created - thus enabling the final model to be more robust. Finally, including training corpora from other Indo-Aryan languages like Urdu or Marathi could be beneficial. Although Subtask 2 focuses on the translation of Hinglish-English, the validation and test sets (as well as training sets) contain many examples of code-mixing between related Indo-Aryan languages and English - most prominently in Urdu, which is historically and linguistically similar to Hindi.

\begin{table*}[!th]
\centering
\begin{tabular}{@{}lcccc@{}}
\toprule
 & BLEU ($\uparrow$) & ChrF++ ($\uparrow$) & TER ($\downarrow$) & WER ($\downarrow$) \\ \midrule
\multicolumn{5}{l}{\textit{Validation Experiments}} \\
AA (dev = MixMT valid) & 20.5 & 41.2 & 72.7 & 78.6 \\
AA (dev = train subset) & 23.3 & 45.7 & 68.3 & 74.6 \\ \midrule
\multicolumn{5}{l}{\textit{Training Experiments (dev=train subset)}} \\
AA (train = all Hg-\textgreater{}En minus XLM BT data) & 18.3 & 38.4 & 78.3 & 83.4 \\
AA (train = all Hg-\textgreater{}En) & 23.3 & 45.7 & 68.3 & \textbf{74.6} \\
AA (train = all Hg-\textgreater{}En + all Hi-\textgreater{}En) & \textbf{24.4} & \textbf{46.2} & \textbf{68.2} & 74.9 \\ \midrule
Random & 24.3 & 45.2 & 68.4 & 74.6 \\ \bottomrule
\end{tabular}
\caption{Aligned Augmentation experiments for subtask 2, as evaluated on the official MixMT Subtask 2 validation set. ``Validation experiments" refers to experiments performed to select the best choice of the validation set for early stopping. `MixMT valid' refers to the same validation set mentioned earlier (that is also used for evaluation), while `train subset' refers to a subset (last 1000 sentences) of the respective training corpus. ``Training experiments'' seek to explore various dataset choices during training time, using a subset from the training corpus for validation. }
\label{tab:aa-results}
\end{table*}

In the end, we observe that the AA models we train are unable to beat our original single-model baseline, despite having more parameters. Curiously, this is also the case for the randomly initialized baseline in Table \ref{tab:aa-results}. Due to time constraints, we are unable to investigate the reasons behind these. Possible explanations could include: training paradigm differences (initializing with hi$\rightarrow$en vs mixing hi$\rightarrow$en with hg$\rightarrow$en), ensembling, experimental setting disparities, inherent differences between training libraries (Fairseq vs Marian). It is possible that addressing these disparities, as well as exploring the directions suggested in the previous paragraph, could enable AA baselines to yield superior results for code-mixed translation.

\begin{table*}[!th]
\centering
\begin{tabular}{@{}ccccccc@{}}
\toprule
 & BLEU & ChrF++ & TER & WER & ROUGE-L & Human Eval. Score \\ \midrule
\textbf{Subtask 1} & 26.9 & 52.7 & 55.2 & 56.2 & 57.9 & 3.85 \\
\textbf{Subtask 2} & 28.7 & 51.2 & 59.1 & 61.3 & 62.5 & 3.75 \\ \bottomrule
\end{tabular}
\caption{Final Test Results for the University of Edinburgh's submissions of MixMT 2022. BLEU, ChrF++ and TER were evaluated by us while WER and ROUGE-L results are from the official Codalab leaderboard. Human evaluation (on a scale of 1-5) was provided by the organisers on 20 random sentences and we report the average.}
\label{tab:test-results}
\end{table*}

\section{Test Results}

The final test results for our submissions are listed in Table \ref{tab:test-results}. For Subtask 1, we used unconstrained decoding with beam-size 12, and for Subtask 2 we used our baseline ensemble (4) with beam-size 36. We evaluated BLEU, ChrF++ and TER ourselves, while the other metrics are provided by the organizers. We ranked second in both subtasks on the MixMT leaderboard\footnote{\href{https://tinyurl.com/codalab-ldbd}{https://tinyurl.com/codalab-ldbd}} although in both the automatic and human evaluation\footnote{\href{https://tinyurl.com/heval-mixmt}{https://tinyurl.com/heval-mixmt}}, there does not appear to be a statistically significant difference. Furthermore, we note that some participants have an exceedingly high number of test submissions and would encourage future shared tasks to put in place measures to avoid this.

%\FloatBarrier % all figures and tables to be rendered before the conclusion.
\section{Conclusion}

In this work, we described our various findings and experiences while building NMT systems that translated between Hinglish and monolingual English/Hindi - as part of the WMT22 Code-Mixing Shared Task. We proposed various corpora that could be useful for these tasks - many of which we create as part of this work - and utilizing these, build high-performing MT systems that, for both subtasks, constituted one of the leading unconstrained models. In addition, we also explored and analysed some alternative approaches for training our models like constrained decoding and Aligned Augmentation (AA) which, despite not beating our original baselines, yielded findings that are useful for future research. Perhaps the most notable of these suggests that efforts to create Hinglish datasets, including using transliterated Hindi as an approximation, can be fruitful and pivotal to high performance. While efforts to handle noise in social media text (such as AA-based pretraining) can also help, further research is required to establish the most optimal ways to do the same.

% \section*{Acknowledgements}
% commented out. we can add it later.

% Entries for the entire Anthology, followed by custom entries

\section{Acknowledgements}

We would like to give special thanks to Nikita Moghe for her valuable feedback and insights throughout this research.

This work was supported in part by the UKRI Centre for Doctoral Training in Natural Language Processing, funded by the UKRI (grant EP/S022481/1) and the University of Edinburgh, School of Informatics. The experiments in this paper were performed using resources provided by the Cambridge Service for Data Driven Discovery (CSD3) operated by the University of Cambridge Research Computing Service\footnote{\href{www.csd3.cam.ac.uk}{www.csd3.cam.ac.uk}}, and using the Sulis Tier 2 HPC platform hosted by the Scientific Computing Research Technology Platform at the University of Warwick. Sulis is funded by EPSRC Grant EP/T022108/1 and the HPC Midlands+ consortium.

\bibliography{anthology, custom}
\bibliographystyle{acl_natbib}

\appendix

\section{XLM details}
\label{sec:xlmrdeets}

In order to backtranslate the Hinglish data, we hoped to train a good quality semi-supervised system using the XLM toolkit \citep{conneau-etal-2020-unsupervised}. We use Masked Language Modelling (MLM) to pretrain a transformer encoder model on English, Hindi and Hinglish monolingual data. The model consisted of 6 layers, 1024 embedding dimensions, batch size 128, and a 0.1 dropout rate. We use 16.5M sentences of English WikiMatrix \citep{schwenk-etal-2021-wikimatrix}, 20M of HindiMono \cite{hindencorp05:lrec:2014} and 18.8M of Hinglish from L3Cube-HingCorpus \citep{nayak2022l3cube} and CC-100 Hindi Romanized \citep{DBLP:conf/acl/ConneauKGCWGGOZ20}. Vocabulary and data preprocessing is the same as for the Marian models (4k BPE merges).

We then initialised a full transformer model with the pretrained encoder, and further trained with denoised autoencoding, MLM, machine translation\footnote{hi$\leftrightarrow$en directions only}, and backtranslation\footnote{Only direction involving hg: hi-hg-hi, en-hg-en, hg-hi-hg, hg-en-hg}objectives. We use the Samanantar corpus (10.1M) for the hi$\leftrightarrow$en translation objective, the 6.7k HinGe sentences as validation for hg$\leftrightarrow$en and hg$\leftrightarrow$hi directions, and the hi-en FLORES development set for hi$\leftrightarrow$en.

\end{document}